\documentclass[sigconf]{acmart}
\usepackage{graphicx}
\usepackage[colorinlistoftodos,prependcaption,textsize=tiny]{todonotes}
\usepackage[utf8]{inputenc}
\usepackage{booktabs}

\begin{document}


\title{Contextual Dialogue Act Classification for Open-Domain Conversational Agents}

\author{Ali Ahmadvand}
\affiliation{\institution{Computer Science Department, \\Emory University, Atlanta, GA}}
\email{ali.ahmadvand@emory.edu}

\author{Jason Ingyu Choi}
\affiliation{\institution{Computer Science Department, \\ Emory University, Atlanta, GA}}
\email{in.gyu.choi@emory.edu}

\author{Eugene Agichtein}
\affiliation{\institution{Computer Science Department, \\ Emory University, Atlanta, GA}}
\email{eugene.agichtein@emory.edu}

\begin{abstract}
\vspace{-1mm}
Classifying the general intent of the user utterance in a conversation, also known as Dialogue Act (DA), e.g., open-ended question, statement of opinion, or request for an opinion, is a key step in Natural Language Understanding (NLU) for conversational agents. While DA classification has been extensively studied in human-human conversations, it has not been sufficiently explored for the emerging open-domain automated conversational agents. Moreover, despite significant advances in utterance-level DA classification, full understanding of dialogue utterances requires conversational context. Another challenge is the lack of available labeled data for open-domain human-machine conversations. To address these problems, we propose a novel method, CDAC (Contextual Dialogue Act Classifier), a simple yet effective deep learning approach for contextual dialogue act classification. Specifically, we use transfer learning to adapt models trained on human-human conversations to predict dialogue acts in human-machine dialogues. To investigate the effectiveness of our method, we train our model on the well-known Switchboard human-human dialogue dataset, and fine-tune it for predicting dialogue acts in human-machine conversation data, collected as part of the Amazon Alexa Prize 2018 competition. The results show that the CDAC model outperforms an utterance-level state of the art baseline by 8.0\% on the Switchboard dataset, and is comparable to the latest reported state-of-the-art contextual DA classification results. Furthermore, our results show that fine-tuning the CDAC model on a small sample of manually labeled human-machine conversations allows CDAC to more accurately predict dialogue acts in real users' conversations, suggesting a promising direction for future improvements.
\end{abstract}

%

\vspace{-0.65cm}
\copyrightyear{2019} 
\acmYear{2019} 
\setcopyright{acmlicensed}
\acmConference[SIGIR '19]{Proceedings of the 42nd International ACM SIGIR Conference on Research and Development in Information Retrieval}{July 21--25, 2019}{Paris, France}
\acmBooktitle{Proceedings of the 42nd International ACM SIGIR Conference on Research and Development in Information Retrieval (SIGIR '19), July 21--25, 2019, Paris, France}
\acmPrice{15.00}

\maketitle
\vspace{-0.37cm}
\section{Introduction and Related Work}
\vspace{-1mm}
\begin{figure*}[htb]
\includegraphics[width=350pt]{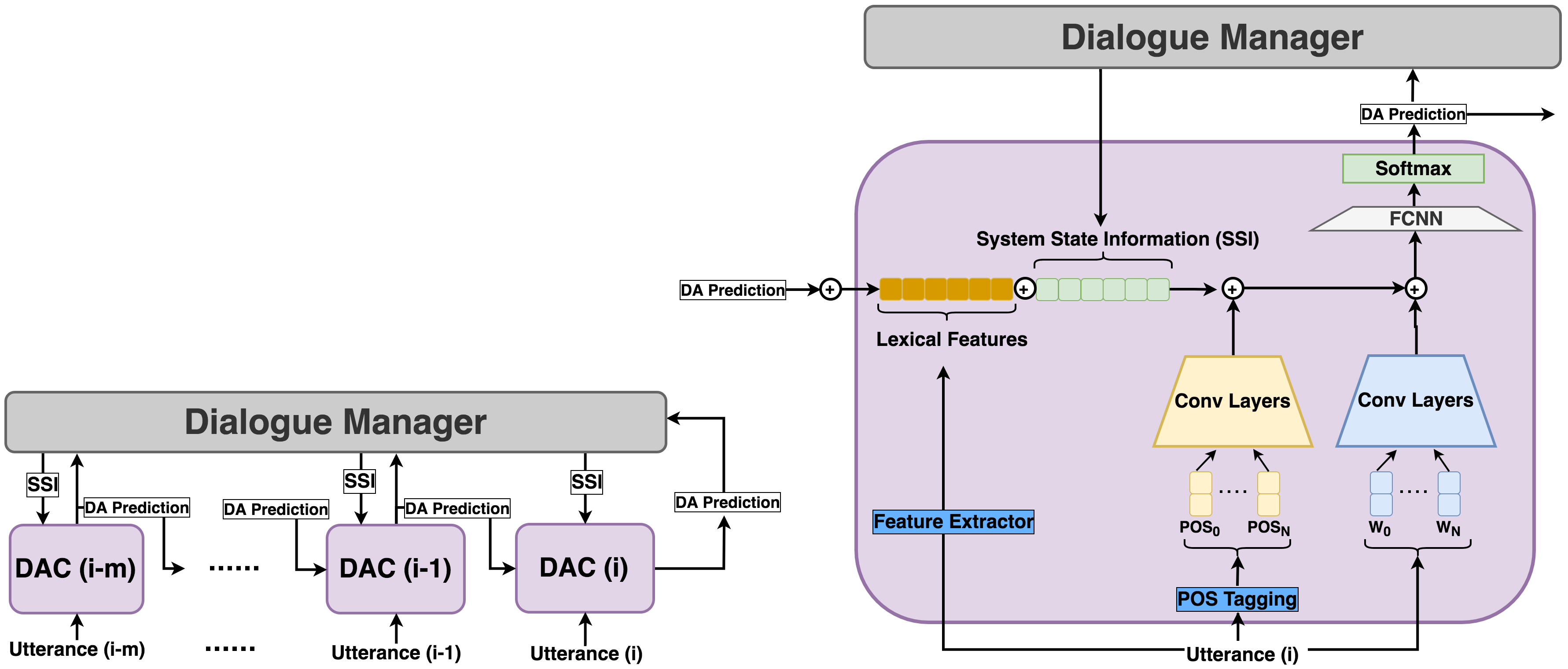}
\caption{Contextual Dialogue Act Classifier (CDAC) architecture overview (left), where ``DAC'' is the dialogue act classifier for individual utterances (shown in detail on right). The extracted features, system state features, and details of the convolution (Conv) layers are described in detail in (Section~\ref{sec:representation}.)}
\label{fig:OverallSystem}
\vspace{-0.5cm}
\end{figure*}

In order for a conversational system to respond properly, it must first understand the intent of the user utterance. Dialogue Act (DA) identification is a traditional approach to intent classification in dialogue system research, which aims to predict the goal of each utterance, such as information request, statement of opinion, greeting, opinion request, and others. ~\cite{DA-Pareti2016} categorized DAs as having two main categories: 1) primary intents; 2) secondary intents. Each of these intents can further be divided into implicit or explicit. Since identifying these intents correctly is crucial for dialogue systems, this problem has been studied extensively for decades, for human-human conversations. Recently, this idea was also extended to human-machine conversations ~\cite{DA-Stolcke2016, DA-Liu2017, DA-Kumar2018}. 

Utterances in natural conversations are contextually dependent in nature, which makes the DA prediction challenging. For example, the utterance like "Oh, yeah" can be interpreted as "Yes-Answer", "Accept-Agree", or "Backchannel", which requires the previous context to disambiguate ~\cite{ DA-Liu2017, DA-Bothe2018}. Therefore, we propose a context-aware model for this task. 

For human-machine conversations, DA classification is more challenging due to three additional factors: 1) Often, human utterances are short (only 2.8 words on average in our data); 2) Automatic Speech Recognition (ASR) is still not quite a human level performance; 3) Lack of available open-domain labeled human-machine conversation data. 
To address these challenges, we propose a deep learning based model, which utilizes contextual evidence, such as preceding utterances alongside the system state information, for more accurate predictions. To reduce requirements for labeled training data, we follow the approach of pre-training the model followed by fine-tuning, which has proven its effectiveness on various natural language processing tasks such as question answering ~\cite{TR-Mou2016, TR-Chung2017}. 

To represent the utterance and system state for each conversation turn, we built on previous studies which identified effective features for human-human DA classification including syntax, prosody, and lexical cues (e.g., ~\cite{DA-Stolcke2016, DA-Grau2004}. 
We integrate many of these ideas into the proposed lexical and syntactic features in our CDAC system, and augment these with representation learning.

Recently, deep learning and representation learning approaches shown promising results on many tasks including text classification and DA classification (e.g., ~\cite{DA-Blunsom2016}). For instance, reference ~\cite{DA-Blunsom2016} proposed a context-based RNN model for dialogue act classification for the human-human Switchboard dataset, while reference ~\cite{DA-Liu2017} proposed a hierarchical CNN and RNN model for this task. 

Reference ~\cite{Khatri:2018} demonstrated the benefits of accurate DA classification for topic classification in open-domain dialogue systems. Inspired by the promising results of ~\cite{Khatri:2018}, ~\cite{Wang:Combining}, and ~\cite{Chen:external}, we propose a novel, yet relatively simple Contextual Dialogue Act Classifier (CDAC) model, which incorporates lexical, syntactic, semantic, and contextual evidence into DA classification. To our knowledge, CDAC is the first to extend and adapt the ideas~\cite{Khatri:2018} for contextual DA classification in open-domain conversational systems, such as those fielded in the Amazon Alexa 2018 Challenge. Interestingly, previous state-of-the-art DA models (e.g., References ~\cite{DA-Liu2017} and ~\cite{DA-Kumar2018}), rely on complete conversations for context, including future utterances, while CDAC uses only the utterances from earlier in the conversation, which makes CDAC feasible for online (real-time) conversational DA classification. As far as we know, CDAC is also the first successful attempt to fine-tune a DA classification model trained on a dataset of {\em human-human conversations}, to predict DAs in open domain {\em human-machine} conversations. 

In summary, our contributions are twofold: (1) development of a novel context-aware Dialogue Classification model, CDAC, for open domain human-machine conversations; (2) demonstrating promising results after fine-tuning CDAC trained on human-human conversations to human-machine conversations, which is a necessary step for intelligent open-domain conversational agents.

\vspace{-0.3cm}
\section{Contextual Dialogue Act Classifier (CDAC) Model}
\label{sec:cdac-model}
In this section, we describe our proposed method, CDAC, for contextual dialogue act classification. First, we describe the features used to represent user utterances, and the conversation and system context. Then, we present the CDAC model architecture, implementation, and training details.

\subsubsection*{\bf \em Content and Context Representation}
\label{sec:representation}
For individual utterance representation, we use both word embedding, and surface (lexical and syntactic) features, described next. The word representation weights are initialized using pre-trained Word2Vec embeddings (300 dimensions), and updated during training. The embedding-based utterance representation is augmented using three types of features: 1) lexical; 2) syntactic; 3) system state information (SSI), summarized below.

\begin{table}[h]
\vspace{-2mm}
\footnotesize
    \begin{tabular}{l|l}
    \toprule
    \bf Lexical Features & \bf Short Description \\
    \bottomrule
    \textit{F}\textsubscript{1} - \textit{Word Count} & Word\textsubscript{count} in utterance \\
    \textit{F}\textsubscript{2} - \textit{Char Count} & Char\textsubscript{count} in utterance \\
    \textit{F}\textsubscript{3} - \textit{Sentence Count} & Sentence\textsubscript{count} in utterance \\
    \textit{F}\textsubscript{4} - \textit{Average Word Count} & Average Word\textsubscript{count} in utterances \\
    \textit{F}\textsubscript{5} - \textit{Average Char Count} & Average Char\textsubscript{count} in utterances \\
    \textit{F}\textsubscript{6} - \textit{IsQuestion} & Binary feature to check for "?" \\

    \toprule
    \bf Syntactic \& SSI Features & \bf Short Description \\
    \bottomrule
    \textit{F}\textsubscript{7} - \textit{POS Tagging} & Part of speech tags\\
    \textit{F}\textsubscript{8} - \textit{Topic Distribution} & Topic distribution vector\\
    \textit{F}\textsubscript{9} - \textit{Suggested Topic}    & Suggested topic to user\\
    \textit{F}\textsubscript{10} - \textit{Suggested Item}    & Suggested entity to user\\
    \textit{F}\textsubscript{11} - \textit{Speaker Id}        & ID assigned to each user\\
    \bottomrule
    \end{tabular}
\label{features}
\vspace{-0.5cm}
\end{table}

The CDAC model combines individual Dialogue Act Classifiers (DACs) for each utterance. Each DAC uses textual features as well as syntactic features, i.e., part of speech tags, are modeled using a convolutional layer pipeline, as shown in Figure~\ref{fig:OverallSystem} (right).
Each ``Conv Layers'' in Figure~\ref{fig:OverallSystem} is a 3-layer CNN with kernels of size 1, 2, and 3, with 100 filter maps for each kernel. Each layer is implemented using a 2D-convolution followed by a max-pooling, batch normalization, and relu activation function. To implement the batch normalization, we used a momentum of $M=0.997$ and an epsilon of $\epsilon=1e-5$.
Then, the output of both pipelines (syntactic and word embeddings) for each utterance representation, and the lexical features, as well as SSI features, are combined through a Fully Connected Neural Network layer (FCNN) with the size of 100. A dropout rate of 0.5 is applied at the FCNN layer to prevent the model from overfitting to the limited training data. Finally, softmax activation is used to obtain the final multi-class DA distribution. For training, categorical cross-entropy loss is minimized using Adam optimizer, with a $\alpha=1e-3$ learning rate and mini-batch size of 64.

The full Contextual DA Classifier model, CDAC, uses the DAC predictions from up to $m$ previous turns in the conversation, where $m$ effectively controls the scope of the conversation context.
These DA predictions for the previous turns are appended to the current utterance representation vector through the FCNN layer, as shown in Figure~\ref{fig:OverallSystem} on right.

The window size $m$ and other hyper-parameters above, largely follow the previous literature. Overall, our CDAC architecture and implementation choices builds on the ideas from the recent state of the art models for DA classification on the Switchboard dataset, while keeping the model simple and inexpensive to train.

\begin{table*}[htb]
    \fontsize{6.0}{6.0}
    \selectfont
    
    \begin{tabular}{p{3.25cm} | p{8.7cm} | p{0.6cm} | p{0.58cm} | p{0.48cm} | p{1.9cm}}
    \toprule
 \textbf{ Utterance} & \textbf{  Response}  & \textbf{  Topic} & \textbf{ PS} & \textbf{ST} & \textbf{DA} \\
 \toprule
 
   Alexa, let's chat  & Nice to meet you! Do you want to talk about music?  & \textit{Music} & \textit{NONE} & \textit{Music} & \textbf{fp: }Opening\\
    I want some recent music from Drake & How about God's plan by Drake?  & \textit{Music} & \textit{Open} & \textit{NONE} & \textbf{sv: }Statement-opinion \\
    I really like that song! & You're welcome. Do you want some more information about Drake? & \textit{Music} & \textit{Music} & \textit{NONE} & \textbf{sv: }Statement-opinion\\ 
    Sure, sounds good & Drake is a Canadian rapper with 30 million followers on Spotify. Do you want some news about him? & \textit{Music} & \textit{Music} & \textit{News} & \textbf{aa: }Accept\\
    I'm good, tell me some fun dog facts & Sure, do you want to know why dogs have wet noses? & \textit{Animal} & \textit{Music} & \textit{NONE} & \textbf{qo: }Open-question \\
    Yes please & Dog's wet noses are for absorbing scent chemicals! Do you want another one? & \textit{Animal} & \textit{Animal} & \textit{NONE} & \textbf{aa: }Accept \\
    Do you have a dog? & Unfortunately, I do not have one. What do you want to chat next? & \textit{Phatic} & \textit{Animal} & \textit{NONE} & \textbf{qy: }Yes-No-Question \\
    Can you tell me about penguins? & Sure, do you want to hear about the fastest penguin species? & \textit{Animal} & \textit{Animal} & \textit{NONE} & \textbf{qo: }Open-Question \\
    Okay, that sounds interesting & Gentoo penguins can reach swimming speed up to 22 mph. Do you want more about penguins? & \textit{Animal} & \textit{Animal} & \textit{NONE} & \textbf{aa: }Accept\\
    No, I'm done chatting for today & Nice to talk to you, good bye! & \textit{Phatic} & \textit{Animal} & \textit{NONE} & \textbf{fc: }Closing
     \\
    \bottomrule
    \end{tabular}
    \caption{An example human-machine conversation, where ``Topic'' is current system state (topic), ``PS'' is the previous system topic, ``ST'' is the system-initiated topic suggestion, if any, and ``DA'' is the manually chosen Dialogue Act label.}
    \label{tab:conversationsample}
\vspace{-0.8cm}
\end {table*}






\subsubsection*{\bf \em System State Information (SSI) Features}
\label{sec:ssi-features}
We hypothesized that modeling context in human-machine conversations is similar to human-human conversations, but with a major difference that the system state information (SSI), unlike the state of a human, can be directly captured and represented as features. We incorporate SSI features alongside the utterance and context representation features for each {\em turn} of the conversation. The SSI features include system topic distribution, the suggested topics (e.g., ``Music''), and suggested items (e.g., specific artists). To encode topic distribution features, we used one-hot encoding, while specific items are represented using Word2Vec word embeddings of the words in the item names.
Note that the SSI features were not used or available for human-human conversations.


\paragraph{\bf \em Transfer Learning from Human-Human Conversations}
To fine-tune the CDAC model from human-human to human-machine conversations, all the weights in the CDAC model are first trained on the human-human Switchboard dataset. Then, all of the network weights are tuned using the Alexa Prize data, but with a smaller learning rate of $\alpha=1e-4$.

\vspace{-0.1cm}
\section{Experimental Setup}
We now introduce the human-human (Switchboard) and human-machine (Alexa Prize) conversation datasets used for training and evaluating CDAC. Then, we explain the human annotation procedure to obtain ground truth labels for the human-machine conversations, followed by the experimental design and metrics. 


\subsubsection*{\bf \em Switchboard Dataset (Human-Human Conversations)}
Switchboard DA corpus ~\cite{Jurafsky1997} is a well-known telephone speech corpus, which contains 42 main DA labels\footnote{\em{\url{https://github.com/cgpotts/swda}}}. The Switchboard corpus contains two official splits: the training split with 1,115 conversations and 196,258 utterances, and the test split, with 40 conversations and 7535 utterances. 

\subsubsection*{\bf \em Alexa Prize 2018 Dataset (Human-Machine Conversations)}
We collected the human-machine conversation data during the Amazon Alexa Prize 2018. 200 conversations of real users with our open-domain conversational agent, containing more than 3,000 utterances were randomly selected. Table \ref{tab:conversationsample} shows an example conversation\footnote{This is a representative conversation between the authors and the real system, since user utterances from live system deployment cannot be reported to protect user privacy.} of a hypothetical (not real) user with the actual system responses. Two different human annotators were asked to manually label two hundred conversations in the human-machine Alexa prize data. The inter-annotator agreement was 0.790, and Kappa was 0.755, indicating strong agreement between the annotators. For the final ground truth label values, in case of disagreements, the label was randomly chosen between the two annotator labels. The distribution of annotated DAs\footnote{See \em \url {http://compprag.christopherpotts.net/swda.html} for full description of DA labels.} is reported in Table \ref{class_dist}. The top four most frequent dialogue acts observed are \underline{Agree/Accept} (aa), \underline{Conventional Opening} (fp), \underline{Reject} (ar), and \underline{Statement Opinion} (sv), accounting for over 68.2\% of the user utterances.

\begin{table}[h]
\footnotesize
      \centering
          \begin{tabular}{l|l||l|l}
          \toprule
            \textbf{DA} & \textbf{Frequency} & \textbf{DA} & \textbf{Frequency}  \\
            \bottomrule
            \textbf{$aa$} & 655 (21.7\%) & \textbf{$fp$} & 501 (16.6\%)\\
            \textbf{$ar$} & 478 (15.8\%) & \textbf{$sv$} & 425 (14.1\%)\\
            \textbf{$qo$} & 227 (7.5\%) & \textbf{$fc$} & 198 (6.6\%)\\
            \textbf{$sd$} & 154 (5.1\%) & \textbf{$b^\wedge m$} & 114 (3.8\%)\\
            \textbf{$no$} & 107 (3.5\%) & \textbf{$qw$} & 80 (2.7\%)\\
            \textbf{$qy$} & 48 (1.6\%) & \textbf{$\%$} & 24 (0.8\%)\\
            \textbf{$ft$} & 7 (0.2\%) & & \\
          \bottomrule
         \end{tabular}
      \caption{Dialogue Acts (DA) frequency distribution in user utterances in the Alexa Prize dataset. }
      \label{class_dist}
\vspace{-0.8cm}
\end{table}

\label{sec:experimental_setup}

\subsubsection*{\bf \em Switchboard Experimental Design}
For CDAC model training, the training conversations were split into 1,000 for training and 115 for validation, leaving the test split untouched during training. For testing, some of the previous studies \cite{DA-Stolcke2016,DA-Blunsom2016,Lee:seq} used only 19 conversations of the available 40 test conversations. Instead, we follow the convention of Liu et al.~\cite{DA-Liu2017} and use all 40 test conversations for evaluation. 
The main baseline model for this experiment is \citep{DA-Stolcke2016}, based on a hidden Markov model, as it remained a state-of-the-art method for more than 10 years. Other state-of-the-art reported results are from the three recent DA classifiers described in references \cite{DA-Bothe2018,DA-Blunsom2016,Lee:seq} respectively.

\subsubsection*{\bf \em Alexa Prize Experimental Design}
For the Alexa prize dataset experiment, Support Vector Machines (SVM) and Multinomial Bayes models are selected as baseline models, using lexical features (words) with tf-idf term weights due to simplicity and low requirements for labeled training data. Contextual features such as previous utterances and system state features are appended to the bag of words feature vector for each utterance. 5-fold cross-validation was used, where 4 folds were used for tuning the weights, and the last fold for the prediction. Finally, following the conventions of the DA classification literature, the main evaluation metric was {\bf overall (micro-averaged) Accuracy}. 

\vspace{-0.1cm}
\section{Results and Discussion}
In this section, we report the overall accuracy of CDAC in comparison to previous state-of-the-art baselines on Switchboard and Alexa data. Feature ablation and error analysis are also reported to provide insights into CDAC system performance. 
\subsubsection*{\bf \em DA Prediction Results on Switchboard Data}
Our main results on the Switchboard dataset are summarized in Table \ref{tab:main}. CDAC improves the baseline model ~\cite{DA-Stolcke2016} by 8.0\%. Moreover, compared to the best known contextual model, we reach comparable results with a more general and simple model. Context window of size 3 yields the strongest performance. However, our results are on all 40 conversations in Switchboard dataset, while ~\cite{DA-Bothe2018} used only 19 of the text conversations. We were unable to replicate the model and results reported in reference ~\cite{DA-Bothe2018}, due to the required model complexity and not having access to a working implementation or sufficient details to reproduce their exact system. In contrast, our proposed model is simpler, while producing state-of-the-art DA classification accuracy on the standard benchmark of human-human conversations. 

\begin{table}[ht]
\begin{center}
\footnotesize

\begin{tabular}{l|c}

\toprule
Methods&   Accuracy  \\ 
\toprule

\textit{Baseline} Stolcke et al.~\cite{DA-Stolcke2016}& 71.00  \\
\midrule
\textit{Previous state of the art methods} & \\
Kalchbrenner et al.~\cite{DA-Blunsom2016} &73.90      \\
Young et al.~\cite{Lee:seq}  & 73.10      \\
\bf Bothe et al.~\cite{DA-Bothe2018}\textbf{*}  &	\bf 77.34 (+8.9\%)    \\
\midrule
CDAC-2 &  76.40    \\
\bf CDAC-3 & \bf 76.70 (+8.0\%)   \\
CDAC-4 & 76.51   \\
\hline
Annotator Agreement & 84.00\\
\hline

\end{tabular}
\end{center}
\caption{ DA classification Accuracy (micro-averaged) on Switchboard dataset, where (*) represents the latest reported state-of-the-art contextual DA classification\cite{DA-Bothe2018}. CDAC-2,-3, and -4 stands for the model using on contexts of size 2, 3 and 4 turns, respectively.}
\label{tab:main}
\vspace{-0.9cm}
\end{table}

\subsubsection*{\bf \em DA Prediction Results on Alexa Prize dataset}
Table \ref{tab:transferlearning} summarizes the baseline and CDAC performances on the human-machine Alexa Prize conversation data. CDAC outperforms traditional classification models such as SVM and Multinomial Bayes (without context) by about 22.7\%. Furthermore, by adding context to SVM and Multinomial Bayes, there are 1.8\% and 4.5\% improvements over the respective baselines. Encouragingly, by pre-training the CDAC model on human-human Switchboard data, and then fine-tuning on the (limited) labeled human-machine data, CDAC achieves an additional 2.7\% improvement. 
It is important to note that despite annotating for only 13 most common DA classes observed
in the human-machine conversation data, compared to the 42 classes in Switchboard human-human data, DA performance degrades on human-machine conversations. This confirms our observation that human-machine DA classification is a more challenging task than for human-human conversations. 

\begin{table}[ht]
\begin{center}
\footnotesize

\begin{tabular}{l|c}

\toprule
Methods&   Accuracy  \\ 
\toprule

\textit{Without context} & \\
Multinomial Bayes & 58.21  \\
SVM & 65.73  \\
\midrule
\textit{With context} & \\
Multinomial Bayes & 59.26  \\
SVM & 68.70  \\
\bf CDAC & \bf 73.25\textbf{*}(+6.6\%)   \\
\bf CDAC + Transfer Learning & \bf 75.34\textbf{*}(+9.6\%)  \\
\hline

\end{tabular}
\end{center}
\caption {DA prediction micro-averaged Accuracy on Alexa dataset with and without context information (size 3 turns), where (*) represents significance levels of p $<$ 0.05.}
\label{tab:transferlearning}
\vspace{-0.9cm}
\end{table}

\subsubsection*{\bf \em Feature Ablation and Error Analysis}
Table \ref{tab:ablation} summarizes the change in accuracy by systematically removing feature sets on the Alexa prize human-machine conversation data. Both lexical and syntactic features are important for DA classification since removing either group decreased the accuracy. 
Interestingly, the most common error is distinguishing between statement-opinion and open-question labels. For instance, the utterance "I like to talk about animals" is challenging to classify, since without context, it is difficult to determine whether a user expressed an opinion, or requested information from the system. Knowing the context and the system state can enable such disambiguation.
\begin{table}[ht]
\footnotesize
\begin{center}
\begin{tabular}{cccc}
\toprule 
Syntactic Features & Lexical Features & Accuracy   \\ 
\toprule
   -      &    -      &   73.94 (-1.64\%)      \\
   -      &    \checkmark     &   74.80 (-0.50\%)      \\
  \checkmark     &    -     &   74.91 (-0.35\%)     \\
   \checkmark     &  \checkmark    &   \bf 75.18     \\
\hline
\end{tabular}
\end{center}
\caption {Feature ablation on CDAC with context window size 1. - and \checkmark indicates features removed and added respectively.}
\label{tab:ablation}
\vspace{-7mm}
\end{table}


In summary, we proposed a contextual Dialogue Act classification model, CDAC, which incorporates lexical, syntactic, and semantic information, in context. Additionally, we introduced a new group of context features to capture the internal system information. Finally, we demonstrated a promising use of fine-tuning on a limited set of labeled human-machine conversations, to decrease manual annotation requirements, and to utilize the existing human-human labeled conversation data. As a result, CDAC was able to outperform state-of-the-art DA classification baselines: by 8.0\% on Switchboard data, and by 9.6\% on the Alexa Data, and performed comparably to the latest reported and more complex state-of-the-art contextual DA classification model. CDAC was also shown to be general enough to be easily fine-tuned for DA classification in human-machine conversations. The implementation is released to the research community\footnote{Available at {\em {\url{https://github.com/emory-irlab/CDAC}}}}.
We believe CDAC represents a promising advance in general user intent classification for intelligent conversational agents.\\

\vspace{-2mm}
\noindent\textbf{Acknowledgments}: We gratefully acknowledge the financial and computing support from the Amazon Alexa Prize 2018.

\vspace{-1mm}
\bibliographystyle{abbrv}
\bibliography{Reference}

\end{document}